%% file: HumanMotionQA.tex
\documentclass[nohyperref]{article}

\usepackage{microtype}
\usepackage{graphicx}
\usepackage{subfigure}
\usepackage{booktabs} 

\usepackage{hyperref}

\usepackage[accepted]{icml2022}

\usepackage{amsmath}
\usepackage{amssymb}
\usepackage{mathtools}
\usepackage{amsthm}

\usepackage[capitalize,noabbrev]{cleveref}
\usepackage{xspace}
\usepackage{xcolor}

\usepackage{subfiles}

\theoremstyle{plain}

\theoremstyle{definition}

\theoremstyle{remark}

\usepackage[textsize=tiny]{todonotes}

\usepackage{multirow}
\newcommand{\bftable}{\fontseries{b}\selectfont}
\newcommand*{\tableindent}{\hspace*{.85cm}}

\definecolor{MyDarkGreen}{rgb}{0.02,0.6,0.02}

\newcommand\dataset{HumanMotionQA\xspace}
\newcommand\model{NSPose\xspace}
\newcommand\datasetname{BABEL-QA\xspace}

\icmltitlerunning{Motion Question Answering via Modular Motion Programs}

\begin{document}

\twocolumn[
\icmltitle{Motion Question Answering via Modular Motion Programs}


\icmlsetsymbol{equal}{*}

\begin{icmlauthorlist}
\icmlauthor{Mark Endo}{equal,cs}
\icmlauthor{Joy Hsu}{equal,cs}
\icmlauthor{Jiaman Li}{cs}
\icmlauthor{Jiajun Wu}{cs}
\end{icmlauthorlist}

\icmlaffiliation{cs}{Department of Computer Science, Stanford University}

\icmlcorrespondingauthor{Mark Endo}{markendo@stanford.edu}

\icmlkeywords{question answering, human motion understanding, neuro-symbolic learning}

\vskip 0.3in
]



\printAffiliationsAndNotice{\icmlEqualContribution} 

\begin{abstract}

In order to build artificial intelligence systems that can perceive and reason with human behavior in the real world, we must first design models that conduct complex spatio-temporal reasoning over motion sequences. Moving towards this goal, we propose the \dataset task to evaluate complex, multi-step reasoning abilities of models on long-form human motion sequences. We generate a dataset of question-answer pairs that require detecting motor cues in small portions of motion sequences, reasoning temporally about when events occur, and querying specific motion attributes. In addition, we propose \model, a neuro-symbolic method for this task that uses symbolic reasoning and a modular design to ground motion through learning motion concepts, attribute neural operators, and temporal relations. We demonstrate the suitability of \model for the \dataset task, outperforming all baseline methods.
\end{abstract}

\section{Introduction}

\subfile{sections/1-introduction}

\section{Related Work}

\subfile{sections/2-related_works}

\section{\dataset and \datasetname}

\subfile{sections/3-task_dataset}

\section{Methods}

\subfile{sections/4-model}

\section{Experiments}

\subfile{sections/5-experiments}

\section{Discussion}

\subfile{sections/6-discussion}

\clearpage

\bibliography{HumanMotionQA}
\bibliographystyle{icml2022}

\newpage
\appendix
\onecolumn

\subfile{sections/7-appendix}

\end{document}

%% file: sections/1-introduction.tex
\begin{figure}[ht!]
\vskip 0.2in
\begin{center}
\centerline{\includegraphics[width=\linewidth]{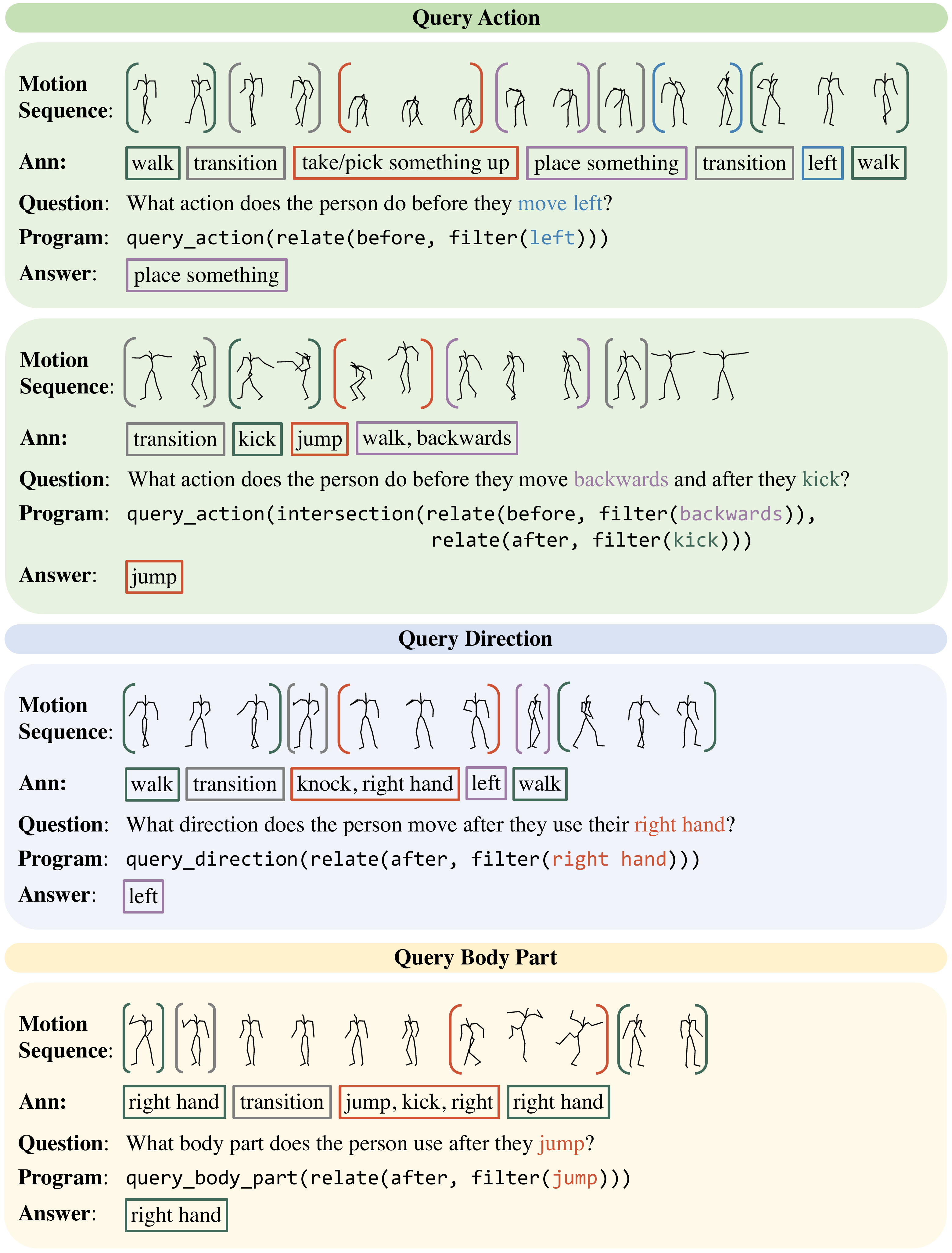}}
\caption{For the task of human motion question answering (\dataset), we create a dataset (\datasetname) that evaluates models' ability to learn complex, multi-step reasoning for human behavior understanding. We present examples of several types of questions in our dataset, including querying for action, direction, and body part across temporal relations.}
\label{pullfigure}
\end{center}
\vskip -0.2in
\end{figure}

A longstanding research goal in artificial intelligence is to build models that can perceive and interact with humans in the real world. To achieve this goal, we must first understand complex human behavior across space and time; hence, we are interested in the characterization of long-form human motion sequences in 3D scenes. The growing amount of available human motion capture data in recent years has enabled the development of a variety of tasks \cite{AMASS, NTU, BABEL}, including action recognition \cite{ActivityNet}, motion forecasting \cite{POTR}, and temporal localization \cite{temporal-localization}. Although these tasks involve the understanding of motion sequences, none require complex, multi-step reasoning about both action-level events (e.g., how behaviors are performed and relate to one another) as well as frame-level fine-grained detection (e.g., body parts involved in specific frames and sudden changes of direction). 

Thus, we propose the task of human motion question answering, \dataset, to evaluate such complex and fine-grained human behavior understanding (See Figure~\ref{pullfigure}). Our task consists of a human motion sequence, paired with a question in natural language and an answer from a vocabulary of words. The questions pertain to different attributes in the motion sequences such as action, direction, and body part, and involve temporal relations such as before, after, and in between. \dataset requires complex motion understanding and spatio-temporal reasoning, as models must (1) detect subtle and complex motor cues performed only in a small portion of a motion sequence and (2) reason temporally about how different sections in a motion sequence relate to one another without having access to action boundaries. To explore the task of HumanMotionQA, we build a dataset \datasetname based on BABEL~\cite{BABEL} and AMASS~\cite{AMASS}. \datasetname comprises 1109 motion sequences with 2577 associated question-answer pairs and is an important step to understanding complex human behavior.

Learning a mapping of human motions and questions to corresponding answers is challenging for two key reasons. First, complex motion reasoning requires grounding different actions in untrimmed motion sequences without access to explicit action boundaries. Second, models typically require large amounts of data and suffer from data biases such as imbalanced action co-occurrences. To enable explicit grounding in untrimmed motion, we propose to decompose the untrimmed sequence into overlapped motion segments so that we can model the relationship between each segment and the question. Moreover, we adopt a neuro-symbolic framework to eliminate the need for large-scale data and mitigate potential data biases. Our proposed approach, \model, executes symbolic programs recursively on the input motion sequence and learns modular motion programs that correspond to different activity classification tasks. Our method jointly learns motion representations and language concept embeddings from motion sequences and question-answer pairs. Compared to end-to-end approaches applied to the \dataset task, \model enables improved temporal grounding capabilities. By leveraging the program structure specified in language, we  achieve effective learning of human motion concepts (e.g. activities such as \textit{walking} and \textit{jumping}, activity characteristics such as \textit{forward} and \textit{backward}, and body parts such as \textit{left arm} and \textit{right leg}), leading to a faithful grounding of human trajectories in motion sequences.

We show that \model results in improved question-answering performance compared to baseline end-to-end methods for the task of \dataset. Our method is capable of complex, multi-step reasoning by using decomposed program structures to learn modular human motion concepts. Importantly, \model learns temporal grounding without action localization supervision, resolving prior neuro-symbolic visual reasoning approaches' need for ground truth segments. In summary, we jointly propose \datasetname, a new dataset for human motion question answering, as well as \model, a neuro-symbolic solution designed for this task. Both extend current deep learning capabilities for human behavior understanding.

%% file: sections/2-related_works.tex
\paragraph{Motion reasoning.}
In recent years, action recognition for human motion has been extensively studied~\cite{yan2018spatial,asghari2020dynamic,caetano2019skelemotion,cai2021jolo,chen2021channel,cheng2020skeleton,choutas2018potion,du2015hierarchical,ke2017new,liu2020disentangling,AGCN,shi2020decoupled}. Leading approaches such as ST-GCN~\cite{yan2018spatial} used a graph convolution model to capture the spatial-temporal relationship among joints in different time steps. A typical research paradigm has been focused on designing robust GCN-based model architectures to improve action recognition accuracy given a sequence of joint positions. Recently, PoseConv3D~\cite{duan2022revisiting} revisited pose representation for the action recognition task and proposed a 3D heatmap volume representation to utilize a powerful 3D-CNN model, leading to superior results compared to previous approaches. Skeleton-based action recognition requires trimmed motion segments as input to estimate the probability of action labels. To predict action labels from untrimmed motion sequences, temporal convolution network~\cite{filtjens2022skeleton, yao2018efficient} and transformer model~\cite{sun2022locate} was adopted to estimate per-frame action probability so that the action localization task can be accomplished by aggregating per-frame predictions. However, these works rely on expensive temporal annotations for action segments and are incapable of providing a fine-grained understanding of long motion sequences that require multi-step reasoning. 
In this work, we aim to ground the actions without the need for temporal action annotations and address the task of human motion question-answering for complex reasoning on human behaviors. 

\paragraph{Joint learning of motion and language.}
Prior work on skeleton-based recognition and localization learned neural models from datasets consisting of paired motion and action labels~\cite{liu2017pku,chereshnev2018hugadb,niemann2020lara}. However, a human motion sequence conveys more than a single action label. We can recognize the moving direction of a walking sequence, perceive the body parts involved in each action and infer the temporal relationships between actions. To provide a detailed description of human motion, recent datasets~\cite{BABEL} annotated natural language on top of the existing motion capture datasets~\cite{AMASS} to facilitate the joint modeling of motion and language. These datasets have led to growing research on generating human motions from language descriptions~\cite{guo2022generating,TEACH,MotionCLIP,petrovich2022temos,zhang2022motiondiffuse,kim2022flame}. For example, conditional VAE was adopted to generate natural human movements conditioned on text~\cite{guo2022generating}. Recently, with the success of the diffusion model in various generative tasks, motion generation results have been greatly improved by applying the diffusion formulation to human motion~\cite{zhang2022motiondiffuse,kim2022flame}. Though the generative task from text has been widely studied based on the datasets with motion and language modalities, the motion recognition and reasoning tasks were neglected in the literature. We propose a motion question-answering task for fine-grained motion understanding in this work.  

\paragraph{Neuro-symbolic approaches.}
Neuro-symbolic approaches have proven to be successful in visual reasoning tasks~\cite{yi2018neural,NSCL}. Neuro-symbolic VQA~\cite{yi2018neural} combined symbolic program execution and visual recognition to address the question-answering task, leading to superior performance in the CLEVR benchmark \cite{johnson2017clevr}. NS-CL~\cite{NSCL} further eliminated the need for dense supervision and designed an effective paradigm to train the neuro-symbolic module by looking at images and reading questions and answers. Recently, neuro-symbolic frameworks have also been extended to temporal reasoning tasks~\cite{chen2021grounding} and 3D reasoning problems~\cite{hong20223d, hsu2023ns3d}, showcasing the capability of grounding concepts with weak supervision and generalizing to new language compositions. Inspired by the success of neuro-symbolic approaches in various tasks, we devise a neuro-symbolic framework for motion sequences to address the task of human motion question-answering with natural supervision (questions and answers). By leveraging paired motion and question-answer pairs, we can ground actions concepts temporally, reason about the temporal relations of action segments, and infer attributes such as the moving direction and the body parts involved in each action.

%% file: sections/3-task_dataset.tex
For the \dataset task, we introduce the \datasetname dataset, which consists of human motion sequences paired with questions in natural language and answers from a vocabulary of words. We describe the task in Section~\ref{task} and the dataset details in Section~\ref{dataset}.

\subsection{The \dataset task}
\label{task}

Given a sequence of human motion capture data represented with 3D joint positions, $\mathcal{S} \in \mathbb{R}^{T\times J\times 3}$, where $T$ is the number of timesteps in the motion sequence and $J$ is the number of joints, and a question about the sequence, the goal of \dataset is to predict the corresponding answer by reasoning about the motion sequence $\mathcal{S}$. Each motion sequence $\mathcal{S}$ consists of a temporal composition of several human actions chained together sequentially. For example, a motion sequence can comprise a person kicking a ball with their left foot, running forward, then jumping. For our task, an example corresponding question is ``What direction does the person move before jumping and after using their left foot?" For a model to reliably answer this question correctly, it must first understand where in the sequence the person is jumping and using their left foot, understand the time period between these two events, and know what direction they are moving in that time frame. Questions in \datasetname require multi-step reasoning -- encompassing human motion classification, attribute-specific queries, and an understanding of temporal relations. 

The \dataset task evaluates how well models can detect subtle motor cues performed on only a portion of long-form motion sequences, and the multi-step reasoning abilities of models to first detect motor cues, then reason temporally about action boundaries, and lastly query attributes relating to actions, direction, and body parts.

\subsection{The \datasetname dataset}
\label{dataset}

To build \datasetname, we create question-answer pairs from motion sequences and annotations in the BABEL dataset \cite{BABEL}. We leverage BABEL, as it contains dense labels that describe each individual action in the temporal composition, in addition to when the action occurs in the motion sequence. This dense information allows us to extract motion concepts from discrete parts of the motion sequences and procedurally build questions by processing temporal relations.

The questions in our dataset relate to three categories of motion attributes: action, direction, and body part. Each attribute contains various concepts such as \textit{walk} and \textit{run} for the action attribute, \textit{forward} and \textit{backward} for the direction attribute, and \textit{right arm} and \textit{left leg} for the body part attribute. To compose questions that require reasoning about these different concepts, we use the following logical building blocks: \texttt{filter}, \texttt{relate}, and \texttt{query}. The \texttt{filter} function selects the subset of motion segments that contain a certain concept. The \texttt{relate} function selects a subset of motion segments that satisfy a certain temporal relation. For example, if you apply a \textit{before} relation to a segment, the function selects the preceding segment. The \texttt{query} function outputs what concept is contained in a motion segment for an attribute of interest.

Our questions follow the structure of first filtering for a concept, optionally applying temporal relation(s), then querying for an attribute. For example, given a sequence of someone throwing a ball with their right hand and then running, we can create 
 a question to first \texttt{filter} for the \textit{run} motion, then add a temporal \texttt{relate} function for the \textit{before} relation, and finally \texttt{query} for the body part. In natural language form, this question is ``What body part does the person use before they run?'' With this question structure, we have three different question types, each categorized by the attribute for the \texttt{query} function. Within each question type, we also categorize sub-question types according to the intermediate relation function (either \textit{before}, \textit{after}, \textit{in between}, or no temporal relation). 

To create question-answer pairs from the BABEL dataset, we first extract motion concepts from the sequences by parsing frame-level label texts and action categories. To avoid creating ambiguous questions, we remove action categories that can contain many different types of movements (e.g., animal behavior). Using these extracted motion concepts with temporal ordering, we then sequentially construct questions with our function building blocks. For each unique concept in the motion sequence (only existing in one segment of the temporal composition), we create new sets of questions by filtering for that segment's concept. We then procedurally generate various types of questions building on this first operation by applying possible temporal relations.

If the segment that immediately precedes the filter segment has extracted motion concepts, then we add a \textit{before} relation and create a query question for each annotated attribute in that segment (e.g., action, direction, and/or body part). Likewise, if the segment that immediately follows the filter segment has an extracted motion concept, then we add an \textit{after} relation and create query questions for each attribute. Note that in the case where the segment immediately preceding or following the filter segment is annotated with the \textit{transition} action, we ignore the segment and look one segment before or after for temporal relations. We can also create questions with both \textit{before} and \textit{after} relations (\textit{in between}) by additionally filtering for a concept for the segment on the other side of the query segment, applying the opposite temporal relation, and combining the two relation outputs with \texttt{intersection} before querying. Lastly, if the filter segment contains additional extracted motion concepts, then we create query questions for each additional attribute without the use of temporal relations. For the BABEL train,  validation, and test splits, we generate every possible question in this format and remove questions with concepts that appear less than eight times. 

\begin{figure}[ht]
\begin{center}
\centerline{\includegraphics[width=\columnwidth]{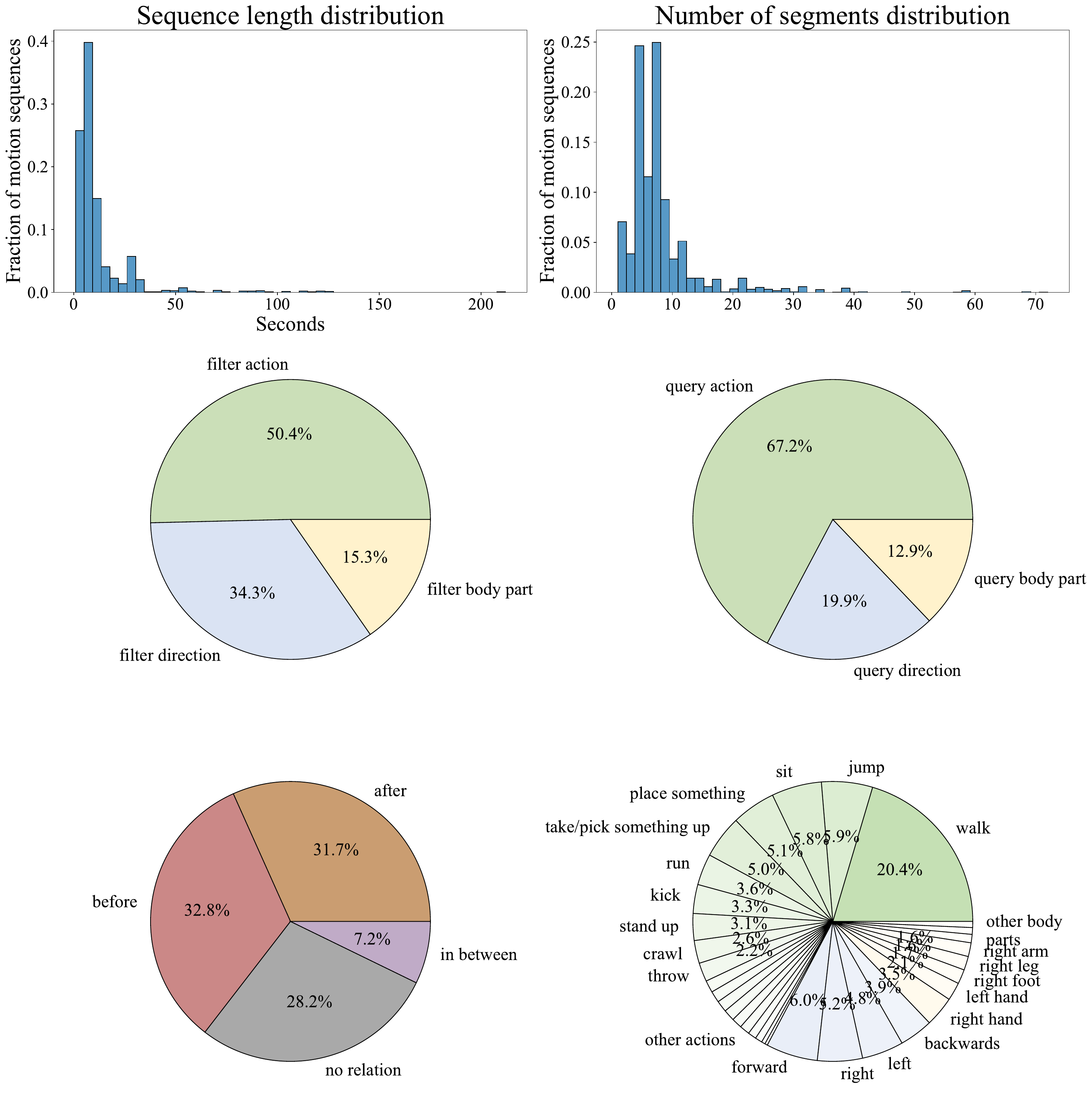}}
\caption{\textbf{Top:} distribution of motion sequence length and number of segments (discrete actions) in motion sequences. \textbf{Bottom:} distribution of filter types, query types, temporal relation types, and query answers}
\label{nsposedatastats}
\end{center}
\vskip -0.2in
\end{figure}

As BABEL consists of natural human motion sequences, certain concepts often occur together either in the same motion segment or adjacent segments. This concurrence of action characteristics causes data bias in co-occurrences between filter concepts and query attribute answers, which systems can easily exploit to answer questions without learning the underlying reasoning process. For example, the answer to the question ``What action does the person do before standing up?'' will often be ``sit down''. To solve this issue, we downsample questions that have common co-occurences. Specifically, given a filter concept $c_i$ and query attribute $a_k$, we count the number of times each answer $\alpha_j$ occurs when first filtering for $c_i$ then querying on $a_k$ (noted as $c_i \rightarrow a_k$). We then balance the dataset such that

$$\frac{\texttt{Count}(\alpha_j)}{\sum_{l \in \text{answers for } c_i \rightarrow a_k} \texttt{Count}(\alpha_l)} < \tau$$

for all $j \in $ answers for $c_i \rightarrow a_k$, where $\tau$ is a threshold set at 34\%.

With this processing, our final dataset is composed of 771 train motion sequences, 167 validation motion sequences, and 171 test motion sequences with an associated 1800 train questions, 384 validation questions, and 393 test questions. Figure~\ref{nsposedatastats} contains information about data statistics. The code for generating this dataset is available at \url{https://github.com/markendo/HumanMotionQA/}. Additional details on the \datasetname dataset and the labeling process can be found in the Appendix.

\begin{figure*}[ht]
\begin{center}
\centerline{\includegraphics[width=\linewidth]{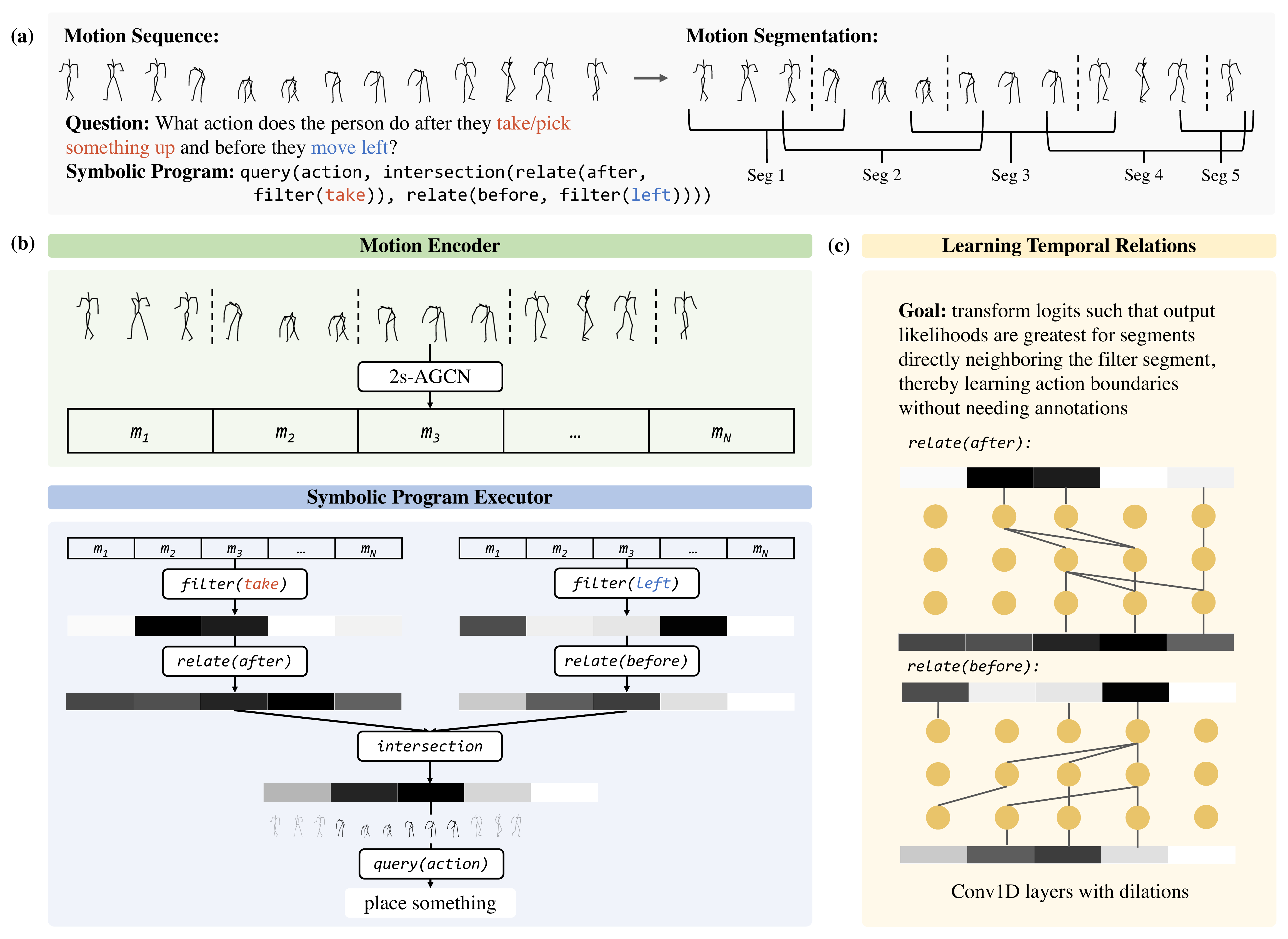}}
\caption{Framework of \model. \textbf{(a)} Overview of extracting motion frames from long-form human motion capture data and the symbolic program structure of \datasetname questions. \textbf{(b)} Visualization of motion feature extraction and program execution using \texttt{filter}, \texttt{relate}, and \texttt{query} functions. \textbf{(c)} Approach of learning relations which are required for the model's temporal understanding and multi-step reasoning abilities.}
\label{nsposearch}
\end{center}
\vskip -0.2in
\end{figure*}

We propose \dataset and \datasetname to evaluate complex reasoning on real-world human motion. As our dataset comes from BABEL, it contains real-world human motion capture of many types of movements. In addition, our dataset is not limited to joint positions as input. \datasetname provides joint position and rotation representations, as well as full body and hand meshes. Importantly, our dataset contains examples sampled from BABEL, which contains different types of actions and a large variation in the composition of motions. Each question is complex and requires reasoning about many aspects of the motion. With these different components, we can test methods' performance on complex real-world reasoning on real-world data.

%% file: sections/4-model.tex
In Section~\ref{main_method}, we present \model, a neuro-symbolic method that we developed to solve the \dataset task. In Section~\ref{baselines}, we discuss additional baselines we explore for question answering in human motion sequences.

\subsection{\model}
\label{main_method}

We introduce \model as a method that leverages a symbolic reasoning process to learn motor cues, modular concepts relating to motion (actions, directions, and body parts), and temporal relations. \model takes as input a human motion sequence as well as an executable program and outputs an answer from a vocabulary of words. We give an overview in Figure~\ref{nsposearch}.

In Figure~\ref{nsposearch} (a), our method first splits the input motion human sequence into $N$ segments. We create overlapping segments of a set frame length such that each segment captures a distinct part of the full sequence with surrounding motion context.

Then, in Figure~\ref{nsposearch} (b), \model{} learns motion encodings for each segment, resulting in modular representations $m_1, ..., m_N$ that span the full motion sequence. Finally, \model{} recursively executes the program trace with motion representations, jointly learning motion concept embeddings and temporal relation transformations. \model{}'s programs are executed as neural networks; in Figure~\ref{nsposearch} (c), the temporal transformation program is implemented as 1D convolutional layers with dilation, enabling learning of temporal action boundaries. The program executor is fully differentiable with respect to the motion representations and concept embeddings, which allows for gradient-based optimization. 

\model improves prior work in neuro-symbolic reasoning in two main ways. The first is the handling of variable length temporal motion sequences, compared to 2D images. We train \model to recognize complex human motion with a skeleton-based feature extraction. The second is \model’s joint learning of action localization and the downstream question answering task. Prior neuro-symbolic visual reasoning approaches such as NS-CL require object-centric input (e.g., object bounding boxes, or translated to our temporal domain, action segments) \cite{NSCL}. \model does this learning jointly through a temporal projection layer, trained in conjunction with the motion feature extractor and the program executor. We detail each part of \model{} below. 

\textbf{Motion feature extractor.}
We use a Two-Stream Adaptive Graph Convolutional Network (2s-AGCN) model to encode motion segments $\mathcal{S}_1, ..., \mathcal{S}_N$ into embedded motion features $m_1, ..., m_N$ \cite{AGCN}. This model goes beyond the conventional GCN approach for skeletal-based action recognition \cite{yan2018spatial} of using a predefined human-body-based graph and instead parameterizes two learned types of graphs. This adaptation increases the flexibility of the model and allows the model to learn different human graph structures for different types of activities. 

Notably, \model operates on full motion sequences, without requiring ground truth action boundaries. We split each input motion sequence into segments of $f$ frames, with varying number of segments in each sequence. We also overlap segments by $o$ frames on each side in order to provide the model with more context in each segment. In our experiments, we set $f = 45$ and $o = 15$. \model's motion feature extractor operates on these frame segmentations, and learns to ground each to a motion concept or attribute. Our method is tasked with action localization in order to answer questions involving temporal operations, while solely supervised by questions and answers in natural language, without pre-training the 2s-AGCN motion encoder.

\textbf{Neuro-symbolic framework.} To answer questions that involve multi-step reasoning about complex activity characteristics across space and time, we propose \model as a neuro-symbolic framework. We extend prior neuro-symbolic visual reasoning methods \cite{NSCL}, which operates on 2D images and requires object segmentations, to \model, which operates on motion sequences and can learn temporal grounding of frames to action concepts without segmentations of action boundaries. We detail \model's program executor below.

First, let us denote $A$ as the set of all motion attributes (e.g., \textit{action}, \textit{direction}, and \textit{body part}) and $C$ as the set of all concepts (e.g., \textit{walk}, \textit{forward}, \textit{left foot}, etc.). For each motion concept $c \in C$, we learn a vector embedding $v^c$ that represents this concept. We also learn an L1-normalized vector $b^c$ that represents the likelihood of the concept belonging to each of the attributes. In addition, we learn neural operators for each attribute $a \in A$ as $u^a$ that transform motion features to the $a$ attribute embedding space.

With these embeddings, vectors, and neural operators, we define the \texttt{filter} and \texttt{query} programs. The \texttt{filter} function takes as input the motion segment embeddings $m_1, ..., m_N$ and a concept of interest $c$ (e.g., \textit{sit}) and returns logits for which segments are most likely to contain the input concept. For a single segment embedding $m_i$, we first calculate the likelihood that $m_i$ includes $c$ as

$$\sigma \left( \sum_{a \in A} \left( b_a^c \cdot \frac{\langle u^a(m_i), v^c \rangle - \gamma}{\tau}\right) \right),$$

where $\sigma$ is the Sigmoid function, $\langle \cdot , \cdot \rangle$ is cosine distance, and $\gamma$ and $\tau$ are scalar constants. In the \texttt{filter} operation, we calculate this likelihood, which we shorten as \texttt{motion\_classify}($m_i$, $c$), for every motion segment.

For the \texttt{query} function, we query an attribute on the motion segments using input segment weights $w_1, ..., w_N$ which are logits returned by either the \texttt{filter} or \texttt{relate} function. We similarly define the likelihood that the input belongs to a concept $c$ as

$$\sum_{i =1}^{N}w_i \cdot \frac{{\texttt{motion\_classify}(m_i, c) \cdot b_a^c}}{\sum_{c' \in C}{\texttt{motion\_classify}(m_i, c') \cdot b_a^{c'}}}.$$

We calculate this likelihood $p_c$ for every concept and define the loss as $-\log\frac{\exp(p_{y})}{\sum_{c \in C}\exp(p_c)}$, where $y$ is the ground truth concept.

\textbf{Temporal grounding.} In addition to learning motion concepts and transformations from the motion to attribute embedding space, we also learn \texttt{relate} operators that capture temporal relations for $\textit{before}$, $\textit{after}$, and $\textit{in between}$ from human motion frames, without the use of annotated action boundaries. The \texttt{relate} functions take in motion segment logits and transform the logits according to the temporal relation of interest, learning action boundaries for the input motion sequence. To learn these temporal transformations, we leverage a convolutional neural network model consisting of 1D convolutional layers with dilation, which has been proven to be successful for learning motifs in sequential data \cite{BPNET}. 

Given the input segment weight vector $W = [w_1, ..., w_N]$ from the preceding filter function, we return CNN($W$), where CNN has three intermediate convolution layers with 16 filters per layer, kernel size of three, and and exponential dilation in every layer. We additionally explore a baseline approach of using a simple linear layer that translates the vector logits to another vector of transformed logits. Though \model is trained with only a final answer cross entropy loss, without any intermediate losses, it is able to learn temporal grounding of frames to action concepts through question answering pairs in natural language as weak supervision.

\model is able to identify boundaries between different actions, as these transition frames are learned implicitly through filtering for concepts in segments with temporal relations. 
We show qualitative results of \model’s temporal grounding capabilities in Figure~\ref{temporal-relation-learning-figure}. Although the predicted boundaries of our model accurately capture transitions, one constraint of these boundaries is that they are predicted at the segment level instead of the model predicting a specific timepoint. To make the boundaries more exact, it is possible to create more segments per motion sequence by reducing the number of frames in each segment. However, the drawback of this change is that there would be less motion context in each segment for the motion encoder to learn from. Through experimentation, we found that having $45$ frame segments with $15$ frames of overlap is a good balance between having large enough segments to learn useful motion cues and having small enough segments to have fine-grain boundary predictions.

\subsection{Baselines}
\label{baselines}

We compare our method against five different baselines. The first baseline uses only question text to answer questions, resulting in a model that can only exploit possible data bias. The second two baselines are built upon a recent method for learning powerful human motion latent representations \cite{MotionCLIP}. The last two baselines are end-to-end methods that leverage question text and the same skeleton-based feature extractor we use in our approach \cite{AGCN}.

\textbf{CLIP.} This method solely uses the question texts and not the motion sequences that are necessary to faithfully answer the corresponding questions. Specifically, we pass the questions into to a pre-trained CLIP model \cite{CLIP} to get text embeddings and then train a simple multilayer perceptron (MLP) on top to predict question answers. We use this method as a rudimentary baseline that can only learn text questions and dataset biases.

\begin{table*}[th!]
\caption{Evaluation of \model and baseline methods on the \datasetname test set. Performance is evaluated using accuracy and we report the mean score of three runs. We find that \model performs better than  baselines. \textsc{Btw} stands for \textit{in between}.}
\label{baselinecomparisontable}
\begin{center}
\begin{scriptsize}
\begin{sc}
\begin{tabular}{l|c|cccc|cccc|cccc}
\toprule

\multirow{2}{*}{Model} & \multirow{2}{*}{Overall} & \multicolumn{4}{c|}{Query Action} & \multicolumn{4}{c|}{Query Direction} & \multicolumn{4}{c}{Query Body Part} \\
& & All & Before & After & Btw & All & Before & After & Btw & All & Before & After & Btw \\ 
\midrule
CLIP & 0.417 & 0.467 & 0.380 & 0.452 & 0.591 & 0.366 & \bftable 0.467 & 0.292 & 0.222 & 0.261 & 0.261 & 0.278 & \bftable 0.333\\
2s-AGCN-MLP & 0.355 & 0.384 & 0.353 & 0.411 & 0.273 & 0.352 & 0.378 & 0.250 & 0.278 & 0.228 & 0.261 & 0.130 & \bftable 0.333\\
2s-AGCN-RNN & 0.357 & 0.396 & 0.349 & 0.396 & 0.409 & 0.352 & 0.400 & 0.396 & 0.278 & 0.194 & 0.261 & 0.111 & 0.167\\
MotionCLIP-MLP & 0.430 & 0.485 & 0.411 & 0.470 & 0.545 & 0.361 & 0.400 & 0.271 & 0.333 & 0.272 & 0.304 & 0.222 & \bftable 0.333\\
MotionCLIP-RNN & 0.420 & 0.489 & 0.461 & 0.441 & 0.606 & 0.310 & 0.400 & 0.333 & 0.222 & 0.250 & \bftable 0.333 & 0.167 & \bftable 0.333\\
NS-Pose (Ours) & \bftable 0.578 & \bftable 0.627 & \bftable 0.618 & \bftable 0.620 & \bftable 0.639 & \bftable 0.598 & 0.389 & \bftable 0.583 & \bftable 0.750 & \bftable 0.325 & 0.296 & \bftable 0.471 & 0.083\\
\bottomrule
\end{tabular}
\end{sc}
\end{scriptsize}
\end{center}
\vskip -0.1in
\end{table*}

\textbf {MotionCLIP-MLP.} In this method, we embed both the natural language questions and motion sequences into the same latent representation space such that the two modalities of data can be easily used together for prediction. To do this, we utilize MotionCLIP, a transformer-based motion auto-encoder trained to reconstruct motion while being aligned to its corresponding text's position in the CLIP space \cite{MotionCLIP}. We pass the entire motion sequence into the model to attain a single motion representation, and we concatenate this information with the CLIP embedding of the question. We then train an MLP on top to predict answers.

\textbf{MotionCLIP-RNN.} For this baseline, we follow a similar setup to MotionCLIP-RNN, except we pass individual action segments into the model instead of the entire motion sequence. This modification results in attaining one representation for each action segment in the sequence. In order to predict the answer, we utilize a recurrent neural network (RNN). Specifically, we first pass the CLIP embedding of the question into the model as the initial hidden state. The latent motion segment representations are then passed sequentially into the model as inputs. We use the final output of the RNN model as the predicted answer to the question. We conjecture that this change from MotionCLIP-MLP to MotionCLIP-RNN will enable this baseline to discern fine-grain details in the motion sequence since each distinct action has its own embedding. The appendix contains visualizations for the MotionCLIP baseline architectures. For both MotionCLIP baselines, we fine-tune the human motion encoder on our dataset while using frozen CLIP weights.

\textbf{2s-AGCN-MLP.} This baseline is an end-to-end approach that leverages 2s-AGCN to extract motion features. 2s-AGCN-MLP uses the same feature extractor as \model, and hence evaluates the importance of modular programs from the symbolic components of \model compared to prior end-to-end regimes. Concatenating a CLIP embedding of the question with a single 2s-AGCN motion representation, we train an MLP on top to predict answers. Similarly to MotionCLIP-MLP, we fine-tune the human motion encoder.

\textbf{2s-AGCN-RNN.} In this setup, we use the same motion feature encoder, 2s-AGCN, but utilize a recurrent neural network (RNN) to predict the answer. We follow the same prediction process as MotionCLIP-RNN but use motion embeddings from 2s-AGCN instead of MotionCLIP.

%% file: sections/5-experiments.tex
\begin{figure*}[ht!]
\vskip 0.2in
\begin{center}
\centerline{\includegraphics[width=\linewidth]{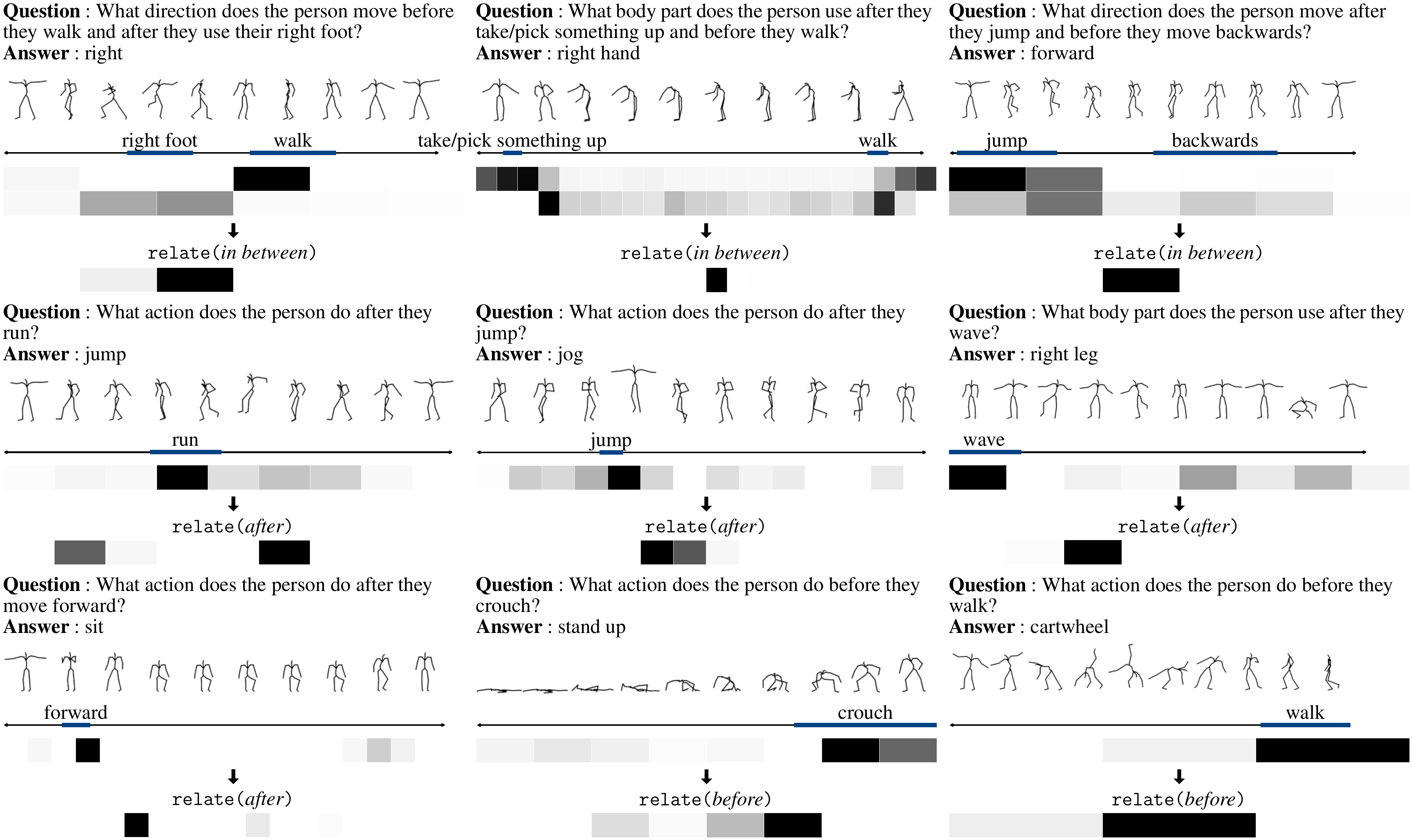}}
\vskip -0.1in
\caption{Visualization of \model's temporal grounding results from weak supervision of question-answer pairs only. For each motion sequence, we present the corresponding question, answer, and motion sequence, along with ground truth action boundaries, as well as predicted boundaries by \model and outputs after temporal \texttt{relate} operators. The rows of rectangles above the \texttt{relate} function represent model outputs for how likely each motion segment contains the filtered concept, where darker-colored squares signify a higher likelihood. The rows of rectangles below the \texttt{relate} function represent model outputs for which segments satisfy the temporal relationship (come directly before, after, or in between the filtered segments). For sequences with two rows of rectangles above the relation function, the two rows represent outputs for the two filtered concepts with the concept appearing first in the question text on top. Note that for visualization purposes, we use the variant of \model without overlapping motion segments.}
\label{temporal-relation-learning-figure}
\end{center}
\vskip -0.3in
\end{figure*}

We investigate the performance of \model and baseline methods on the \datasetname test set. Table \ref{baselinecomparisontable} contains detailed results of all methods. We compare \model to baseline methods in Section~\ref{baselines_res} and present ablations of \model in Section~\ref{ablation}.

\subsection{Comparison to baselines}
\label{baselines_res}

Our findings show that \model outperforms all the baseline methods in overall accuracy. Notably, our method outperforms the deeper MotionCLIP baselines, which are pre-trained on the BABEL dataset to learn CLIP-aligned human motion latent representations. Our method has an overall performance improvement over CLIP by 0.161, an improvement over MotionCLIP-MLP by 0.148, and an improvement over MotionCLIP-RNN by 0.158. \model also significantly outperforms both end-to-end 2s-AGCN baselines. The relatively low performance of 2s-AGCN-MLP and 2s-AGCN-RNN shows that our method does not owe its success to the 2s-AGCN motion feature extractor.

We conjecture that the improved performance of \model is due to our neuro-symbolic approach that learns modular programs. Instead of exploiting data bias during training, our model learns to ground individual motion concepts and can be accurately applied to the validation set with different compositions of motion concepts. We present full results comparing different methods in Table \ref{baselinecomparisontable}.

\subsection{Ablation studies}
\label{ablation}

\begin{table*}[ht!]
\caption{Ablations of \model{} on the \datasetname test set. Performance is evaluated using accuracy and we report the mean score of three runs. \textsc{Qu.} stands for \texttt{query} and \textsc{Fil.} stands for \texttt{filter}.}
\label{ablations-table}
\begin{center}
\begin{scriptsize}
\begin{sc}
\begin{tabular}{l|l|c|cc|cc|cc}
\toprule
Seg Strat & Temp Ground & Overall & Qu. Action & Fil. Action & Qu. Direction & Fil. Direction & Qu. Body Part & Fil. Body Part\\
\midrule
$f$ + $o$ & Conv1D & 0.578 & 0.627 & 0.509 & 0.598 & 0.473 & 0.325 & 0.454\\
$f$ & Conv1D & 0.540 & 0.573 & 0.457 & 0.577 & 0.463 & 0.332 & 0.424\\
$f$ & Linear & 0.540 & 0.602 & 0.505 & 0.548 & 0.529 & 0.275 & 0.495\\
\midrule
GT & Conv1D & 0.553 & 0.601 & 0.620 & 0.583 & 0.505 & 0.271 & 0.313\\
GT & Linear & 0.549 & 0.606 & 0.626 & 0.587 & 0.599 & 0.266 & 0.460\\
\bottomrule
\end{tabular}
\end{sc}
\end{scriptsize}
\end{center}
\vskip -0.2in
\end{table*}

We also show ablations with different setups of \model. In Section~\ref{motionsegstrat}, we compare our method of splitting motion sequences into segments of $f$ frames with $o$ frames of segment overlap, to the approach of not overlapping frames, and a variant that leverages ground truth action boundary annotations to create motion segments. In Section~\ref{learningtemp}, we examine different temporal relation functions. Table \ref{ablations-table} contains the results of the various setups.

\subsubsection{Motion segmentation strategy}
\label{motionsegstrat}

We compare \model's weakly-supervised approach of grounding temporal action compositions through segmenting motion sequences into $n$ frame segments with $o$ frames of overlap to (1) a simpler approach without frame overlap, and (2) the more annotation-intensive approach of using ground truth action annotations for creating motion segments (See Table~\ref{ablations-table}).

We find that the frame overlapping approach has an overall performance improvement of 0.038 over the method without frame overlap. We hypothesize that overlapping segments add important motion context for improving representations from the feature extractor while maintaining fine-grain information that comes from having a large number of segments.

Overall, we find that our weakly-supervised approach outperforms the variant of \model using ground truth action boundaries by 0.025. This performance difference demonstrates that \model can faithfully and accurately reason about complex human behavior across time from full motion sequences. See Figure~\ref{temporal-relation-learning-figure} for examples of \model's program execution for temporal relations. We provide additional analyses on \model performance in the Appendix.

\subsubsection{Temporal relation function}
\label{learningtemp}

We present ablations for two different strategies of learning temporal relations. We show experiment results from leveraging our proposed model consisting of 1D convolutional layers with dilation, and experiment results using a simple linear model, for the temporal operator. We find that the convolutional approach has similar overall accuracy as the linear approach. The similar performance between the two methods shows that our framework can accurately learn temporal relation transformations using simple functions.

%% file: sections/6-discussion.tex
In this work, we propose the task of human motion question answering, \dataset, for human behavior understanding, and propose \model as a neuro-symbolic solution for this task. \dataset evaluates models' ability to conduct complex and fine-grained multi-step reasoning across subtle motor cues in motion sequences. \model approaches this task by decomposing questions into program structures that are executed recursively on the input motion sequence, and learns modular programs that correspond to different activity classification tasks. Our method exhibits fine-grain reasoning abilities about complex motions and learns temporal grounding from question answering, leading to improved human behavior understanding.

A limitation of \model is its dependency on pre-defined motion programs instead of using a semantic parser to translate natural language questions into programs. We do not learn semantic parsing from text, as our focus is on the temporal grounding of motion sequences. A future direction is the inclusion of a trained semantic parsing module to translate questions into programs, enabling broader applicability of our method.

\paragraph{Acknowledgments.}

We thank Sumith Kulal for providing valuable feedback on the paper. This work is in part supported by Stanford Institute for Human-Centered Artificial Intelligence (HAI), Stanford Wu Tsai Human Performance Alliance, Toyota Research Institute (TRI), NSF RI \#2211258, ONR MURI N00014-22-1-2740, AFOSR YIP FA9550-23-1-0127, Analog Devices, JPMorgan Chase, Meta, and Salesforce.

%% file: sections/7-appendix.tex
\section{Supplementary for Motion Question Answering via Modular Motion Programs}

\subsection{Domain-specific language \& program implementations}

We define the domain-specific language (DSL) used for the \dataset task. Table \ref{programdefinitions} includes signatures and semantics for all functions, and Table \ref{functionimplementations} includes implementations for all functions.

\begin{table}[h]
\caption{Operations used in the programs of \dataset.}
\label{programdefinitions}
\begin{center}
\begin{tabular}{lll}
\toprule

Function & Signature & Semantics \\
\midrule
\texttt{Sequence} & () $\rightarrow$ SegmentSet & Return all motion segments in the sequence. \\
\midrule
\texttt{Filter} & (SegmentSet, Concept) $\rightarrow$ SegmentSet & Filter for motion segments that contain a concept. \\
\midrule
\texttt{Relate} & (SegmentSet, Relation) $\rightarrow$ SegmentSet & Outputs segments that satisfy the temporal relationship. \\
\midrule
\texttt{Query} & (SegmentSet, Attribute) $\rightarrow$ Concept & Queries the attribute of the SegmentSet. \\
\midrule
\texttt{Intersection} & (SegmentSet, SegmentSet) $\rightarrow$ SegmentSet & Outputs the intersection of the two segment sets. \\
\bottomrule
\end{tabular}
\end{center}
\vskip -0.1in
\end{table}

\begin{table}[h]
\caption{Implementations for all functions used in the programs of \dataset.}
\label{functionimplementations}
\begin{center}
\begin{tabular}{ll}
\toprule

Signature & Implementation \\
\midrule
$\texttt{Sequence}() \rightarrow y: \text{SegmentSet}$ & $y_i = 10$, for all $i \in \{1, ..., N\}$\\
\midrule
$\texttt{Filter}(x: \text{SegmentSet}, c: \text{Concept}) \rightarrow$ & $y_i = \min (x_i, \texttt{motion\_classify}(m_i, c))$ \\
$\tableindent y: \text{SegmentSet}$ &\\
\midrule
$\texttt{Relate}(x: \text{SegmentSet}, rel: \text{Relation}) \rightarrow$ & $y = \text{Linear}_{rel} (x)$ or $y = \text{CNN} _{rel}(x)$ \\
$\tableindent y: \text{SegmentSet}$ & \\
\midrule
$\texttt{Query}(x: \text{SegmentSet}, a: \text{Attribute}) \rightarrow$ & $y = \arg\max_{c \in C}\left(\sum_{i = 1}^{N} x_i \cdot \frac{\texttt{motion\_classify}(m_i, c) \cdot b^c_a}{\sum _{c' \in C}\texttt{motion\_classify}(m_i, c') \cdot b^{c'}_a}\right)$ \\
$\tableindent y: \text{Concept}$ & \\
\midrule
$\texttt{Intersection}(x: \text{SegmentSet},$ & $z_i = \min (x_i, y_i)$ \\
$\tableindent y: \text{SegmentSet}) \rightarrow z: \text{SegmentSet}$ & \\
\bottomrule
\end{tabular}
\end{center}
\vskip -0.1in
\end{table}

\clearpage

\subsection{Full results}
We report the complete results of all methods and setup for each of three runs in Table \ref{allruninfo}.

\begin{table*}[th!]
\caption{Evaluation of various \model setups and baseline methods on the \datasetname test set. Accuracy is reported for all runs.}
\label{allruninfo}
\begin{center}
\begin{scriptsize}
\begin{sc}
\begin{tabular}{l|cccc|cccc|cccc}
\toprule

\multirow{2}{*}{Model} & \multicolumn{4}{c|}{Query Action} & \multicolumn{4}{c|}{Query Direction} & \multicolumn{4}{c}{Query Body Part} \\
& All & Before & After & Btw & All & Before & After & Btw & All & Before & After & Btw \\ 
\midrule
\multirow{3}{*}{CLIP} & 0.456 & 0.349 & 0.433 & 0.591 & 0.389 & 0.467 & 0.375 & 0.333 & 0.267 & 0.304 & 0.278 & 0.250\\
 & 0.487 & 0.395 & 0.478 & 0.636 & 0.333 & 0.467 & 0.188 & 0.167 & 0.267 & 0.261 & 0.278 & 0.500\\
 & 0.460 & 0.395 & 0.444 & 0.545 & 0.375 & 0.467 & 0.312 & 0.167 & 0.250 & 0.217 & 0.278 & 0.250\\
 \midrule
\multirow{3}{*}{2s-AGCN-MLP} & 0.418 & 0.360 & 0.422 & 0.318 & 0.361 & 0.467 & 0.125 & 0.500 & 0.200 & 0.261 & 0.056 & 0.500\\
 & 0.398 & 0.407 & 0.444 & 0.318 & 0.319 & 0.200 & 0.312 & 0.167 & 0.183 & 0.174 & 0.167 & 0.250\\
 & 0.337 & 0.291 & 0.367 & 0.182 & 0.375 & 0.467 & 0.312 & 0.167 & 0.300 & 0.348 & 0.167 & 0.250\\
 \midrule
\multirow{3}{*}{2s-AGCN-RNN} & 0.372 & 0.314 & 0.400 & 0.500 & 0.403 & 0.467 & 0.375 & 0.333 & 0.200 & 0.261 & 0.111 & 0.250\\
 & 0.456 & 0.419 & 0.467 & 0.455 & 0.306 & 0.467 & 0.375 & 0.167 & 0.233 & 0.304 & 0.167 & 0.250\\
 & 0.360 & 0.314 & 0.322 & 0.273 & 0.347 & 0.267 & 0.438 & 0.333 & 0.150 & 0.217 & 0.056 & 0.000\\
 \midrule
\multirow{3}{*}{MotionCLIP-MLP} & 0.487 & 0.430 & 0.478 & 0.455 & 0.361 & 0.333 & 0.312 & 0.333 & 0.250 & 0.217 & 0.278 & 0.250\\
 & 0.498 & 0.407 & 0.500 & 0.591 & 0.361 & 0.467 & 0.250 & 0.333 & 0.267 & 0.304 & 0.222 & 0.250\\
 & 0.471 & 0.395 & 0.433 & 0.591 & 0.361 & 0.400 & 0.250 & 0.333 & 0.300 & 0.391 & 0.167 & 0.500\\
 \midrule
\multirow{3}{*}{MotionCLIP-RNN} & 0.490 & 0.453 & 0.456 & 0.636 & 0.375 & 0.533 & 0.375 & 0.167 & 0.233 & 0.261 & 0.167 & 0.500\\
 & 0.502 & 0.453 & 0.444 & 0.591 & 0.236 & 0.333 & 0.188 & 0.167 & 0.267 & 0.348 & 0.222 & 0.000\\
 & 0.475 & 0.477 & 0.422 & 0.591 & 0.319 & 0.333 & 0.438 & 0.333 & 0.250 & 0.391 & 0.111 & 0.500\\
 \midrule
\multirow{3}{*}{NS-Pose ($f$ + $o$, Conv1D)} & 0.633 & 0.629 & 0.648 & 0.667 & 0.603 & 0.417 & 0.607 & 0.750 & 0.321 & 0.306 & 0.529 & 0.000\\
 & 0.611 & 0.589 & 0.592 & 0.750 & 0.603 & 0.375 & 0.679 & 0.750 & 0.355 & 0.306 & 0.412 & 0.250\\
 & 0.636 & 0.637 & 0.620 & 0.500 & 0.590 & 0.375 & 0.464 & 0.750 & 0.299 & 0.278 & 0.471 & 0.000\\
 \midrule
\multirow{3}{*}{NS-Pose ($f$, Conv1D)} & 0.579 & 0.540 & 0.570 & 0.472 & 0.581 & 0.375 & 0.536 & 0.750 & 0.359 & 0.389 & 0.353 & 0.500\\
 & 0.578 & 0.556 & 0.563 & 0.528 & 0.587 & 0.500 & 0.357 & 0.750 & 0.363 & 0.333 & 0.176 & 0.250\\
 & 0.561 & 0.540 & 0.542 & 0.444 & 0.565 & 0.542 & 0.429 & 0.750 & 0.274 & 0.389 & 0.059 & 0.250\\
 \midrule
\multirow{3}{*}{NS-Pose ($f$, Linear)} & 0.622 & 0.621 & 0.556 & 0.472 & 0.609 & 0.375 & 0.500 & 0.750 & 0.167 & 0.278 & 0.118 & 0.000\\
 & 0.593 & 0.589 & 0.528 & 0.528 & 0.590 & 0.375 & 0.429 & 0.500 & 0.338 & 0.278 & 0.235 & 0.500\\
 & 0.591 & 0.565 & 0.507 & 0.528 & 0.446 & 0.125 & 0.214 & 0.250 & 0.321 & 0.417 & 0.176 & 0.500\\
 \midrule
\multirow{3}{*}{NS-Pose (GT, Conv1D)} & 0.637 & 0.661 & 0.634 & 0.556 & 0.596 & 0.500 & 0.357 & 0.750 & 0.226 & 0.333 & 0.059 & 0.250\\
 & 0.584 & 0.597 & 0.563 & 0.583 & 0.565 & 0.458 & 0.429 & 0.250 & 0.316 & 0.333 & 0.353 & 0.250\\
 & 0.581 & 0.532 & 0.620 & 0.583 & 0.587 & 0.458 & 0.464 & 0.750 & 0.269 & 0.278 & 0.176 & 0.250\\
 \midrule
\multirow{3}{*}{NS-Pose (GT, Linear)} & 0.611 & 0.589 & 0.627 & 0.750 & 0.593 & 0.333 & 0.464 & 0.750 & 0.192 & 0.194 & 0.265 & 0.000\\
 & 0.596 & 0.556 & 0.592 & 0.417 & 0.531 & 0.375 & 0.143 & 0.250 & 0.335 & 0.306 & 0.324 & 0.500\\
 & 0.611 & 0.573 & 0.599 & 0.722 & 0.637 & 0.500 & 0.393 & 0.750 & 0.269 & 0.194 & 0.412 & 0.000\\
\bottomrule
\end{tabular}
\end{sc}
\end{scriptsize}
\end{center}
\vskip -0.1in
\end{table*}

\subsection{Failure mode analyses}
We note some areas where models may fail to answer questions from our dataset correctly. One such failure case is when sequences have transition frames between the filter segment and the segment being queried on. For example, in one question the person is moving to the right for $15$ frames, transitioning for $20$ frames, using their left hand for $22$ frames, transitioning for $18$ frames, then moving forward for $94$ frames. The associated question is “what body part does the person use after they move right and before they move forward?” The periods of transition from one action to the next make the temporal relations less reliable, which will ultimately make the segment weights inaccurate for the query function. Another difficulty with this question is that the person is only using their left hand for $22$ frames, which is a very small portion of the overall motion sequence. With the transition periods making the temporal relations difficult and the preciseness needed to pinpoint a body part used in only $22$ frames, models are not able to answer this type of question with high accuracy.

We additionally hypothesize that the low performance on query body part questions with between relations is partly due to the fact that encoded motion features don’t capture information about body parts very well. Without sufficient information about body location in the embeddings, the learned neural operator for body parts will be ineffective and the transformation from motion features to the body part embedding space will therefore be unreliable. This is supported by the fact that querying body parts is the question type with lowest accuracy across methods. 

\clearpage

\begin{figure}[ht]
\vskip 0.2in
\begin{center}
\centerline{\includegraphics[width=\columnwidth]{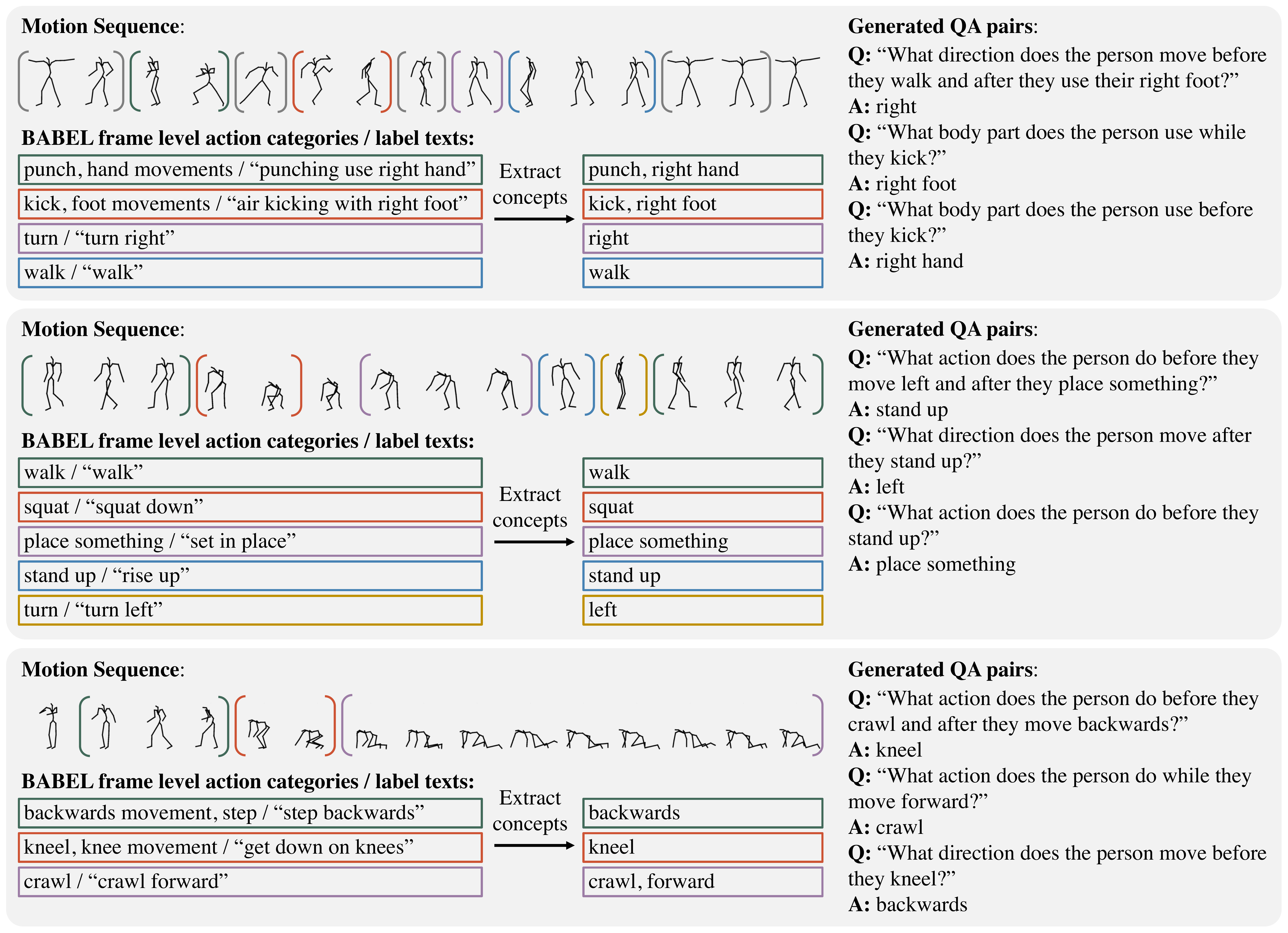}}
\caption{Qualitative examples of extracting motion concepts from BABEL labels and generating question-answer pairs for \datasetname. Motion segments in \textcolor{gray}{gray} boxes are annotated with the \textit{transition} action in BABEL.}
\label{labeling-figure}
\end{center}
\vskip -0.2in
\end{figure}

\subsection{\dataset and \datasetname}
Our \dataset task and \datasetname dataset differ from existing video question-answering datasets in two key ways. First, while existing video QA datasets cover reasoning with actions, we aim to address a more fine-grained human behavior understanding problem (for example, what body part is involved in each action). Second, our dataset lies in a different domain of skeleton-based human motion instead of third-person view videos. Such datasets that consist of skeleton-based human motion and corresponding diverse, natural language question-answer pairs do not previously exist.

The benefits of using skeleton representation are as follows. First, as discussed in previous work on skeleton-based action recognition and localization \cite{xu2022skeleton, sun2022locate}, skeleton-based representation eliminates the nuisances of 2D videos such as lighting changes, background variations, etc, and the 3D joint representation is a more compact human-centric representation. Second, skeleton-based representation can be applied in various applications where videos are not convenient to capture. For example, our skeleton-based neuro-symbolic framework can be generalized to analyze 3D human motion reconstructed from different modalities, for example, motion reconstructed from sparse IMU sensors \cite{jiang2022transformer} or egocentric videos \cite{luo2021dynamics}, which enable applications in analyzing everyday activities of people or monitoring actions of physical impaired people (where motions are usually reconstructed from egocentric signal).

In addition, although we categorize our questions into three types, our dataset provides coverage across a variety of aspects of human motion. First, the questions within each question type are diverse. Within each question type, there are numerous motion concepts that can be filtered for, and temporal relations add an additional element of complexity and variation. Second, the motion sequences have a large variation in terms of types of movements, lengths of sequences, duration of actions, and compositions of different movements. With that said, there is a significant amount of questions pertaining to querying actions, as it is a key temporal feature in motion sequences. We built \datasetname from the original real-world dataset, where annotators were asked to write descriptions of the motion sequences, which includes naming all the actions in the video.

\subsection{Labeling process}
The \datasetname labels for frame-level texts and action categories are provided by the BABEL dataset. They were originally collected by showing videos of motion sequences from AMASS to human annotators. The human annotators described a list of actions performed in the motion sequences and delineated start and end times from each of the described actions. From these raw frame-level texts, the authors clustered the labels to map them to a set list of action categories. More information about this process can be found in section 3.4 of the BABEL paper.

In our work, we extract motion concepts by parsing these frame-level label texts and action categories. For actions, we extract non-ambiguous action categories. For body parts and direction, we search through the label texts and extract concepts that are written in the texts. As an example, given the action category / label text pairs of (punch, ``punching use right hand"), (kick / foot movements, ``air kicking with right foot"), (turn , ``turn right"), and (walk, ``walk"), from the first segment we can extract \textit{punch} and \textit{right hand} concepts, from the second segment we can extract \textit{kick} and \textit{right foot} concepts, from the third segment we can extract the \textit{right} concept, and from the fourth segment we can extract the \textit{walk} concept. Figure \ref{labeling-figure} contains qualitative examples of the data creation process.

\clearpage
\subsection{Baseline information}
\begin{figure}[ht]
\vskip 0.2in
\begin{center}
\centerline{\includegraphics[width=\columnwidth]{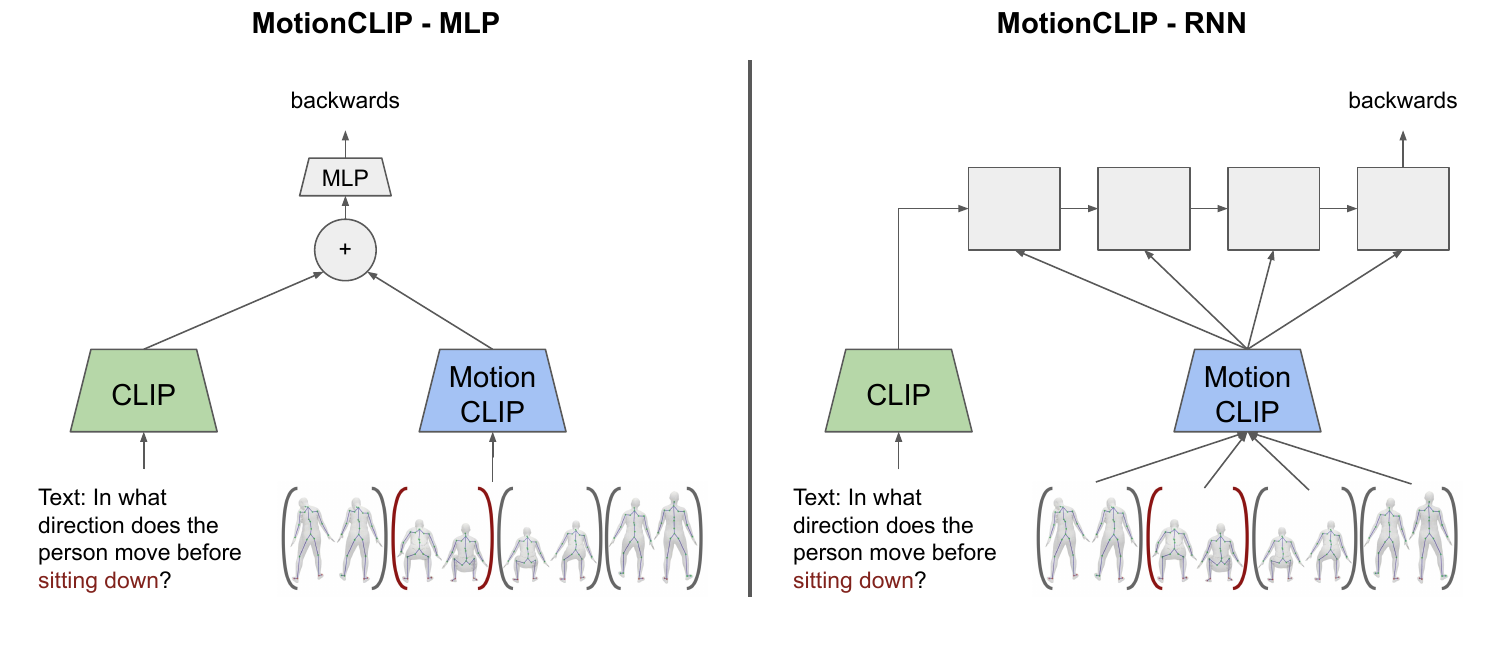}}
\caption{Visualizations of the MotionCLIP-MLP (left side) and MotionCLUP-RNN (right side) baseline methods.}
\label{baseline-figure}
\end{center}
\vskip -0.2in
\end{figure}

We visualize the differences between the MotionCLIP-MLP and MotionCLIP-RNN approaches in Figure \ref{baseline-figure}.
